\RequirePackage{fix-cm}
\documentclass[twocolumn]{svjour3}      
\smartqed  
\usepackage{graphicx}
\usepackage[numbers,sort&compress]{natbib}

\usepackage{booktabs}
\usepackage{siunitx}
\usepackage{mathtools}

\usepackage{tikz} 
\usepackage{multirow} 
\usepackage{graphicx} 

\usepackage{subfig}
\usepackage{amsmath}
\usepackage{url}

\date{}

\usepackage{xcolor}

\usepackage[keeplastbox]{flushend}
\usepackage[a4paper=false,
            citecolor=blue,
            colorlinks=true,
            urlcolor=blue,
            linkcolor=blue,
            pdfauthor={Germain Forestier},
            pdftitle={Accurate and interpretable evaluation of surgical skills from kinematic data using fully convolutional neural networks},
            pdfsubject={Accurate and interpretable evaluation of surgical skills from kinematic data using fully convolutional neural networks},
            pdfkeywords={kinematic data, surgical education, deep learning, time series classification, interpretable machine learning&}
            ]{hyperref}
\usepackage{textcomp}
\usepackage{tikz}
\newcommand\copyrighttext{%
\small This is the author's version of an article published in the International Journal of Computer Assisted Radiology and Surgery.
The final authenticated version is available online at: \url{https://doi.org/10.1007/s11548-019-02039-4}.}
\newcommand\copyrightnotice{%
\begin{tikzpicture}[remember picture,overlay]
\node[anchor=north,yshift=-40pt,xshift=-9pt] at (current page.north) {\fbox{\parbox{\dimexpr\textwidth-\fboxsep-\fboxrule\relax}{\copyrighttext}}};
\end{tikzpicture}
}

\begin{document}

\title{Accurate and interpretable evaluation of surgical skills from kinematic data using fully convolutional neural networks}

\titlerunning{Deep learning for surgical skill evaluation}        

\author{Hassan Ismail Fawaz         \and
        Germain Forestier  \and 
        Jonathan Weber  \and 
        Lhassane Idoumghar \and 
        Pierre-Alain Muller
}


\institute{   IRIMAS, Universit\'e Haute Alsace \\
              12 Rue des Fr\`eres Lumière,\\
              68093 Mulhouse, France\\
              Tel.: +33-3-89336960\\
              \email{\{hassan.ismail-fawaz,germain.forestier,\\ jonathan.weber,lhassane.idoumghar,pierre-alain.muller\}@uha.fr}\\
}


\maketitle

\begin{abstract}~

\noindent
\textit{Purpose}
Manual feedback from senior surgeons observing less experienced trainees is a laborious task that is very expensive, time-consuming and prone to subjectivity.
With the number of surgical procedures increasing annually, there is an unprecedented need to provide an accurate, objective and automatic evaluation of trainees' surgical skills in order to improve surgical practice.\\
\textit{Methods}
In this paper, we designed a convolutional neural network (CNN) to classify surgical skills by extracting latent patterns in the trainees' motions performed during robotic surgery.  
The method is validated on the JIGSAWS dataset for two surgical skills evaluation tasks: classification and regression.\\
\textit{Results}
Our results show that deep neural networks constitute robust machine learning models that are able to reach new competitive state-of-the-art performance on the JIGSAWS dataset.  
While we leveraged from CNNs' efficiency, we were able to minimize its black-box effect using the class activation map technique.\\
\textit{Conclusions}
This characteristic allowed our method to automatically pinpoint which parts of the surgery influenced the skill evaluation the most, thus allowing us to explain a surgical skill classification and provide surgeons with a novel personalized feedback technique.
We believe this type of interpretable machine learning model could integrate within ``Operation Room 2.0'' and support novice surgeons in improving their skills to eventually become experts.
\end{abstract}

\copyrightnotice{}

\section{Introduction}

Over the last century, the standard training exercise of Dr. William Halsted has dominated surgical education in various regions of the world~\citep{polavarapu2013100}.
His training methodology of ``see one, do one, teach one'' is still one of the most adopted approaches to date~\citep{ahmidi2017a}.
The main idea is that the student could become an experienced surgeon by observing and participating in mentored surgeries~\citep{polavarapu2013100}.
These training techniques, although widely used, lack of an objective surgical skill evaluation method~\citep{kassahun2016surgical}. 
Standard assessment of surgical skills is presently based on checklists that are filled by an expert watching the surgical task~\citep{ahmidi2017a}.
In an attempt to predict a trainee's skill level without using on an expert surgeon's judgement, objective structured assessment of technical skills (OSATS) was proposed and is currently adopted for clinical practice~\citep{niitsu2013using}.
Alas, this type of observational rating still suffers from several external and subjective factors such as the inter-rater reliability, the development process and the bias of respectively the checklist and the evaluator~\citep{hatala2015constructing}. 

Further studies demonstrated that a vivid relationship occurs between a surgeon's technical skill and the postoperative outcomes~\citep{bridgewater2003surgeon}. 
The latter approach suffers from the fact that the aftermath of a surgery hinges on the physiological attributes of the patient~\citep{kassahun2016surgical}. 
Furthermore, obtaining this type of data is very strenuous, which renders these skill evaluation techniques difficult to carry out for surgical education.
Recent progress in surgical robotics such as the \emph{da Vinci} surgical system~\cite{davinci} enabled the recording of video and kinematic data from various surgical tasks.
Ergo, a substitute for checklists and outcome-based approaches is to generate, from these kinematics, global movement features (GMFs) such as the surgical task's speed, time completion, motion smoothness, curvature and other holistic characteristics~\citep{zia2017automated,kassahun2016surgical}. 
While most of these techniques are efficacious, it is not perspicuous how they could be leveraged to support the trainee with a detailed and constructive feedback, in order to go beyond a naive classification into a skill level (i.e., expert, intermediate, etc.). 
This is problematic as feedback on medical practice enables surgeons to reach higher skill levels while improving their performance~\cite{islam2016affordable}. 

\begin{sloppypar}
Lately, a field entitled \emph{Surgical Data Science}~\citep{maier-hein2017surgical} has emerged by dint of the increasing access to a huge amount of complex data which pertain to the staff, the patient and sensors for capturing the procedure and patient related data such as kinematic variables and images~\citep{gao2014jhu}.
Instead of extracting GMFs, recent inquiries have a tendency to break down surgical tasks into finer segments called ``gestures'', manually before training the model, and finally estimate the trainees' performance based on their assessment during these individual gestures~\citep{lingling2012sparse}. 
Even though these methods achieved promising and accurate results in terms of evaluating surgical skills, they necessitate labeling a huge amount of gestures before training the estimator~\citep{lingling2012sparse}.
We pointed out two major limits in the actual existing techniques that estimate surgeons' skill level from their corresponding kinematic variables: firstly, the absence of an interpretable result of the skill prediction that can be used by the trainees to reach higher surgical skill levels; secondly, the requirement of gesture boundaries that are pre-defined by annotators which is prone to inter-annotator reliability and time-consuming~\citep{vedula2016analysis}. 
\end{sloppypar}

In this paper, we design a novel architecture of convolutional neural networks (CNNs) dedicated to evaluating surgical skills.
By employing one-dimensional kernels over the kinematic time series, we avoid the need to extract unreliable and sensitive gesture boundaries.
The original hierarchical structure of our model allows us to capture global information specific to the surgical skill level, as well as to represent the gestures in latent low-level features.
Furthermore, to provide an interpretable feedback, instead of using a dense layer like most traditional deep learning architectures~\citep{zhou2016learning}, we place a global average pooling (GAP) layer which allows us to take advantage from the class activation map (CAM), proposed originally by~\cite{zhou2016learning}, to localize which fraction of the trial impacted the model's decision when evaluating the skill level of a surgeon.
Using a standard experimental setup on the largest public dataset for robotic surgical data analysis: the JHU-ISI Gesture and Skill Assessment Working Set (JIGSAWS)~\cite{gao2014jhu}, we show the precision of our FCN model. 
Our main contribution is to demonstrate that deep learning can be leveraged to understand the complex and latent structures when classifying surgical skills and predicting the OSATS score of a surgery, especially since there is still much to be learned on what does exactly constitute a surgical skill~\citep{kassahun2016surgical}.

\section{Background}\label{sec:back}

In this section, we turn our attention to the recent advances leveraging the kinematic data for surgical skills evaluation.
The problem we are interested in requires an input that consists of a set of time series recorded by the da Vinci's motion sensors representing the input surgery and the targeted task is to attribute a skill level to the surgeon performing a trial.
One of the earliest work focused on extracting GMFs from kinematic variables and training off-the-shelf classifiers to output the corresponding surgical skill level~\citep{kassahun2016surgical}.
Although these methods yielded impressive results, their accuracy depends highly on the quality of the extracted features. 
As an alternative to GMF-based techniques, recent studies tend to break down surgical tasks into smaller segments called surgical gestures, manually before the training phase, and assess the skill level of the surgeons based on their fine-grained performance during the surgical gestures, for example, using sparse hidden Markov model (S-HMM)~\citep{lingling2012sparse}. 
Although the latter technique yields high accuracy, it requires manual segmentation of the surgical trial into fine-grained gestures, which is considered expensive and time-consuming. 
Hence, recent surgical skills evaluation techniques have focused on algorithms that do not require this type of annotation and are mainly data driven~\citep{IsmailFawaz2018evaluating,zia2017automated,wang2018deep,forestier2017discovering}.
For surgical skill evaluation, we distinguish two tasks. 
The first one is to output the discrete skill level of a surgeon such as novice (N), intermediate (I) or expert (E). 
For example, \cite{zia2017automated} adopted the approximate entropy (ApEn) algorithm to extract features from each trial which are later fed to a nearest neighbor classifier.
More recently,~\cite{wang2018deep} proposed a CNN-based approach to classify sliding windows of time series; therefore, instead of outputting the class for the whole surgery, the network is trained to output the class in an online setting for each window. 
In~\cite{forestier2018surgical}, the authors emphasized the lack of explainability for these latter approaches, by highlighting the fact that interpretable feedback to the trainees is important for a novice to become an expert surgeon~\citep{islam2016affordable}. 
Therefore, the authors proposed an approach that uses a sliding window technique with a discretization method that transforms the time series into a bag of words and trains a nearest neighbor classifier coupled with the cosine similarity.
Then, using the weight of each word, the algorithm is able to provide a degree of contribution for each sliding window and therefore give some sort of useful feedback to the trainees that explains the decision taken by the classifier. 
Although the latter technique showed interesting results, the authors did sacrifice the accuracy in favor of interpretability. 
On the other hand, using our fully convolutional neural networks (FCN) we provide the trainee with an interpretable yet very accurate model by leveraging the class activation map (CAM) algorithm, originally proposed for computer vision tasks by~\cite{zhou2016learning}.
The second type of problem in surgical skill evaluation is to train a model that predicts the modified OSATS score for a certain surgical trial.
For example, \cite{zia2017automated} extended their ApEn model to predict the OSATS score, also known as global rating score (GRS). 
Interestingly, the latter extension to a regression model instead of a classification one enabled the authors to propose a technique that provides interpretability of the model's decision, whereas our neural network provides an explanation for both classification and regression tasks. 

We present briefly the dataset used in this paper as we rely on the features' definitions to describe our method.
The JIGSAWS dataset, first published by~\cite{gao2014jhu}, has been collected from eight right-handed subjects with three different surgical skill levels: novice (N), intermediate (I) and expert (E), with each group having reported, respectively, less than 10 h, between 10 and 100 h and more than 100 h of training on the Da Vinci. 
Each subject performed five trials of each one of the three surgical tasks: suturing, needle passing and knot tying.  
For each trial, the video and kinematic variables were registered.
In this paper, we focused solely on the kinematics which are numeric variables of four manipulators: right and left masters (controlled by the subject's hands) and right and left slaves (controlled indirectly by the subject via the master manipulators). 
These 76 kinematic variables are recorded at a frequency of 30 Hz for each surgical trial.  
Finally, we should mention that in addition to the three self-proclaimed skill levels (N,I,E), JIGSAWS also contains the modified OSATS score~\citep{gao2014jhu}, which corresponds to an expert surgeon observing the surgical trial and annotating the performance of the trainee. 
The main goal of this work is to evaluate surgical skills by considering either the self-proclaimed discrete skill level (classification) or the OSATS score (regression) as our target variable.
We conceive each trial as a multivariate time series (MTS) and designed a one-dimensional CNN dedicated to learn automatically useful features for surgical skill evaluation in an end-to-end manner~\cite{IsmailFawaz2018deep}.

\begin{figure*}
    \centering
    \includegraphics[width=.8\linewidth]{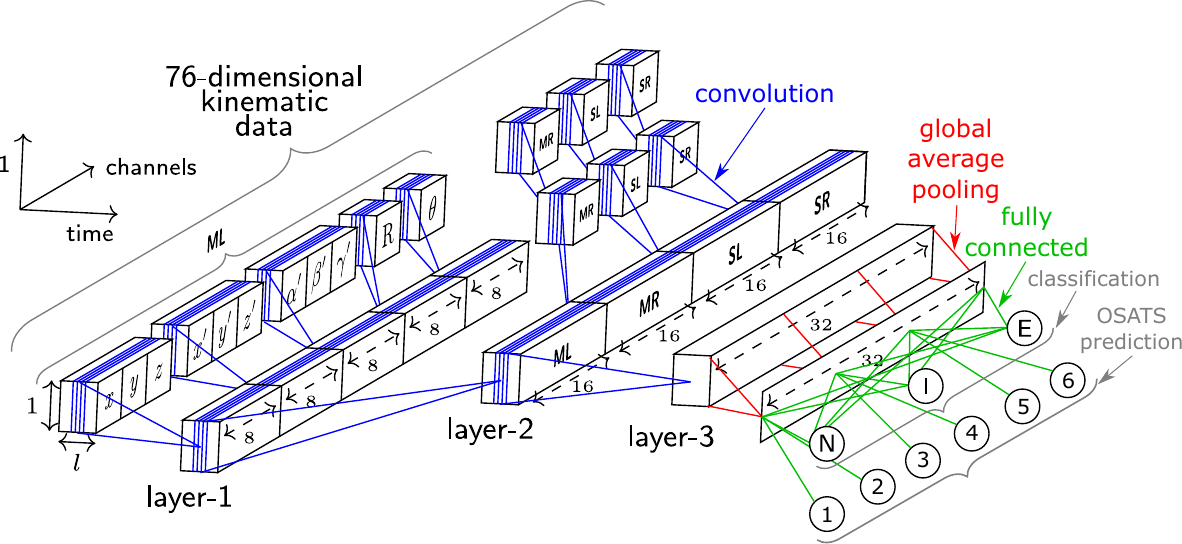}
    \caption{Fully convolutional network (FCN) for surgical skill evaluation.}
    \label{fig:archi}
\end{figure*}

\section{Methods}\label{sec:method}

Our approach takes inspiration of the recent success of CNNs for time series classification~\citep{wang2017time,IsmailFawaz2018deep}. 
Figure~\ref{fig:archi} illustrates the fully convolutional neural network (FCN) architecture, which we have designed specifically for surgical skill evaluation using temporal kinematic data.
The network's input is an MTS with a variable length $l$ and 76 channels. 
For the classification task, the output layer contains a number of neurons equal to three (N,I,E) with the softmax activation function, whereas for the regression task (predicting the OSATS score), the number of neurons in the last layer is equal to six: 
(1) ``Respect for tissue'';
(2) ``Suture/needle handling'';
(3) ``Time and motion'';
(4) ``Flow of operation'';
(5) ``Overall performance'';
(6) ``Quality of final product''~\citep{gao2014jhu}, with a linear activation function.

Compared with convolutions for image recognition, where usually the model's input exhibits two spatial dimensions (height and width) and three channels (red, green and blue), the input to our network is a time series with one spatial dimension (surgical task's length $l$) and 76 channels (denoting the 76 kinematics: $x,y,z,\dots$).
One of the main challenges we have encountered when designing our architecture was the large number of channels (76) compared to the traditional red, green and blue channels (3) for the image recognition problem. 
Hence, instead of applying the filters over the whole 76 channels at once, we propose to carry out different convolutions for each group and subgroup of channels.
We used domain knowledge when grouping the different channels, in order to decide which channels should be clustered together.

Firstly, we separate the 76 channels into four distinct groups, such as each group should contain the channels from one of the manipulators: the first, second, third and fourth groups correspond to the four manipulators (ML: master left, MR: master right, SL: slave left and SR: slave right) of the \emph{da Vinci} surgical system. 
Thus, each group assembles $19$ of the total kinematic variables.  
Next, each group of 19 channels is divided into five different subgroups each containing variables that we believe should be semantically clustered together.
For each cluster, the variables are grouped into five sub-clusters: 
\begin{itemize}
    \item First sub-cluster with three variables for the Cartesian coordinates ($x,y,z$);
    \item Second sub-cluster with three variables for the linear velocity ($x^{\prime},y^{\prime},z^{\prime}$);
    \item Third sub-cluster with three variables for the rotational velocity ($\alpha^{\prime},\beta^{\prime},\gamma^{\prime}$);
    \item Fourth sub-cluster with nine variables for the rotation matrix R;
    \item Fifth sub-cluster with one variable for the gripper angular velocity ($\theta$).  
\end{itemize}
Figure~\ref{fig:archi} illustrates how the convolutions in the first layer are different for each subgroup of kinematic variables. 
Following the same line of thinking, the convolutions in the second layer are different for each group of variables (SL, SR, ML and MR). 
However, in the third layer, the same filters are applied for all dimensions (or channels), which corresponds to the traditional CNN. 

To take advantage from the CAM method while reducing the number of parameters (weights) in our network, we employed a global average pooling operation after the last convolutional layer. 
In other words, the convolution's output (the MTS) will shrink from a length $l$ to 1, while maintaining the same number of dimensions in the third layer.
Without any sort of validation, we choose the following default hyperparameters.
We used $8$ kernels for the first convolution, and then we doubled the number of kernels, thus allowing us to balance the number of parameters for each layer as a function of its depth.
We used ReLU as the nonlinear hidden activation function for all convolutional layers with a stride of $1$ and a kernel length equal to $3$. 

We fixed our objective loss function to be the categorical cross-entropy to learn the network's parameters in an end-to-end manner for the classification task, and the mean squared error (MSE) when learning a regressor to predict the OSATS score, which can be written as: 
\begin{equation}
    \mathrm{MSE}=\frac{1}{n}\sum_{i=1}^{n}(Y_i-\hat{Y}_i)^2.
\end{equation}
The network's weights were optimized using the Adam optimization algorithm~\citep{kingma2015adam}. 
The default value of the learning rate was fixed to $0.001$ as well as the first and second moment estimates were set to $0.9$ and $0.999$ respectively. 
We initialized the weights using Glorot's uniform initialization~\citep{glorot2010understanding}.
The network's parameters were updated with back-propagation using stochastic gradient descent. 
We randomly shuffled the training set before each epoch, whose maximum number was set to $1000$ epochs.
We then saved the model at each training iteration by choosing the network's state that minimizes the loss function on a random (non-seen) split from the training data.
This process is also referred to as ``model checkpoint'' by the deep learning community~\citep{chollet2015keras}, allowing us to choose the best number of epochs based on the validation loss.
Finally, to avoid overfitting, we added an $l2$ regularization parameter whose default value was fixed to $10^{-5}$; however, similarly to the learning rate, we further discuss the effect of this hyperparameter in Sect.~\ref{sec:res}. 
For each surgical task, we have trained a different network, resulting in three different models.\footnote{\scriptsize Our source code is available at \url{https://github.com/hfawaz/ijcars19}}
We adopted for both classification and regression tasks a leave-one-super-trial-out (LOSO) scheme~\cite{ahmidi2017a}.

The use of a GAP layer allows us to employ the CAM algorithm, which was originally designed for image classification tasks by~\cite{zhou2016learning} and later introduced for time series data in~\cite{wang2017time}. 
Using the latter technique, we are able to highlight which fractions of the surgical trial contributed highly to the classification.
Let $A_k(t)$ be the result of the third convolution which is an MTS with $K$ dimensions (here $K$ is equal to 32 filters (by default) and $t$ denotes the time dimension). 
Let $w_k^c$ be the weight between the output neuron of class $c$ and the $k^{th}$ filter. 
Since a GAP layer is used, the input to the output neuron of class $c$ can be written as $z_c$ and the CAM as $M_c$: 
\begin{equation}
\begin{aligned}\label{eq-1}
z_c=\sum_k{w_k^c\sum_t{A_k(t)}}=\sum_t{\sum_k{w_k^c A_k(t)}}\quad , \\ \quad  M_c(t)=\sum_k{w_k^c A_k(t).}
\end{aligned}
\end{equation}
$M_c(t)$ denotes the contribution of each time stamp $t$ when identifying a class $c$. 
Finally, for the regression task, the CAM can be extended in a trivial manner: Instead of computing the contribution to a classification, we are computing the contribution to a certain score prediction (1 out of 6 in total).

\begin{table*}
	\centering
\setlength\tabcolsep{4.5pt}
	\begin{tabular}{l|ccc|ccc|ccc}
		\toprule
		\multirow{2}{*}{\scriptsize Method} &
		\multicolumn{3}{c}{\scriptsize Suturing} &
		\multicolumn{3}{c}{\scriptsize Needle passing} &
		\multicolumn{3}{c}{\scriptsize Knot tying} \\
		& {\scriptsize Micro} & {\scriptsize Macro} & {\scriptsize $\rho$} & {\scriptsize Micro} & {\scriptsize Macro} & {\scriptsize $\rho$} & {\scriptsize Micro} & {\scriptsize Macro} & {\scriptsize $\rho$} \\
		\midrule
		S-HMM~\citep{lingling2012sparse} & 97.4 & n/a & n/a  & 96.2 & n/a& n/a & 94.4 & n/a& n/a \\
		ApEn~\citep{zia2017automated} & \textbf{100} & n/a & 0.59 & \textbf{100} & n/a & 0.45 & \textbf{99.9} & n/a & \textbf{0.66}\\
		Sax-Vsm~\citep{forestier2017discovering} & 89.7 & 86.7& n/a & 96.3 & 95.8& n/a & 61.1 & 53.3& n/a \\
		CNN~\citep{wang2018deep} & 93.4 & n/a& n/a & 89.9 & n/a& n/a & 84.9 & n/a& n/a \\
		FCN (proposed) & \textbf{100} & \textbf{100} & \textbf{0.60} & \textbf{100} & \textbf{100} &\textbf{0.57} & 92.1 & \textbf{93.2} &0.65 \\
		\bottomrule
	\end{tabular}
	\caption{Micro, macro and Spearman's coefficient $\rho$ for surgical skill evaluation.}
    \label{tab:res-class}
\end{table*}

\section{Results}\label{sec:res}

The first task, which we have originally tackled in~\cite{IsmailFawaz2018evaluating}, consists in assigning a skill level for an input surgical trial out of the three possible levels: novice (N), intermediate (I) and expert (E).
In order to compare with current state-of-the-art techniques, we adopted the \emph{micro} and \emph{macro} measures defined in~\cite{ahmidi2017a}.
The \emph{micro} measure refers simply to the traditional \emph{accuracy} metric. 
However, the \emph{macro} takes into consideration the support of each class in the dataset, which boils down to computing the \emph{precision} metric. 
Table~\ref{tab:res-class} reports the \emph{macro} and \emph{micro} metrics of five different models for the surgical skill classification of the three tasks: suturing, knot tying and needle passing. 
For the proposed FCN model, we average the accuracy over 40 runs to reduce the bias induced by the randomness of the optimization algorithm. 
From these results, it appears that FCN is much more accurate than the other approaches with 100\% accuracy for the needle passing and suturing tasks.
As for the knot tying task, we report 92.1\% and 93.2\%, respectively, for the \emph{micro} and \emph{macro} configurations.
When comparing the other four techniques, for the knot tying surgical task, FCN exhibits relatively lower accuracy, which can be explained by the minor difference between the experts and intermediates for this task: Mean OSATS score is 17.7 and 17.1 for expert and intermediate, respectively.  

A sparse hidden Markov model (S-HMM) was designed to classify surgical skills~\cite{lingling2012sparse}.
Although this approach does leverage the gesture boundaries for training purposes, our method is much more accurate without the need to manually segment each surgical trial into finer gestures. 
\cite{zia2017automated} introduced approximate Entropy (ApEn) to generate characteristics from each surgical task, which are later given to a classical nearest neighbor classifier with a cosine similarity metric. 
Although ApEn and FCN achieved state-of-the-art results with 100\% accuracy for the first two surgical tasks, it is still not obvious how ApEn could be used to give feedback for the trainee after finishing his/her training session.
\cite{forestier2017discovering} introduced a sliding window technique with a discretization method to transform the MTS into bag of words.
To justify their low accuracy, the authors in~\cite{forestier2017discovering} insisted on the need to provide \emph{explainable} surgical skill evaluation for the trainees. 
On the other hand, FCN is equally \emph{interpretable} yet much more accurate; in other words, we do not sacrifice accuracy for interpretability. 
Finally,~\cite{wang2018deep} designed a CNN whose architecture is dependent on the length of the input time series.
This technique was clearly outperformed by our model which reached better accuracy by removing the need to pre-process time series into equal length thanks to the use of GAP.

In this paper, we extend the application of our FCN model~\citep{IsmailFawaz2018evaluating} to the regression task: predicting the OSATS score for a given input time series. 
Although the community made a huge effort toward standardizing the comparison between different surgical skills evaluation techniques~\citep{ahmidi2017a}, we did not find any consensus over which evaluation metric should be adopted when comparing different regression models. 
However,~\cite{zia2017automated} proposed the use of Spearman's correlation coefficient (denoted by $\rho$) to compare their 11 combination of regression models. 
The latter is a nonparametric measure of rank correlation that evaluates how well the relationship between two distributions can be described by a monotonic function.
In fact, the regression task requires predicting six target variables; therefore, we compute $\rho$ for each target and finally report the corresponding mean over the six predictions.
By adopting the same validation methodology proposed by~\cite{zia2017automated}, we are able to compare our proposed FCN model to their best performing method. 
Table~\ref{tab:res-class} reports also the $\rho$ values for the three tasks, showing how FCN reaches higher $\rho$ values for two out of three tasks. 
In other words, the prediction and the ground truth OSATS score are more correlated when using FCN than the ApEn-based solution proposed by~\cite{zia2017automated} for the second task and equally correlated for the other two tasks.

\begin{figure*}
    \centering
    \subfloat[Suturing task for an expert]{
    
 \includegraphics[width=.35\linewidth]{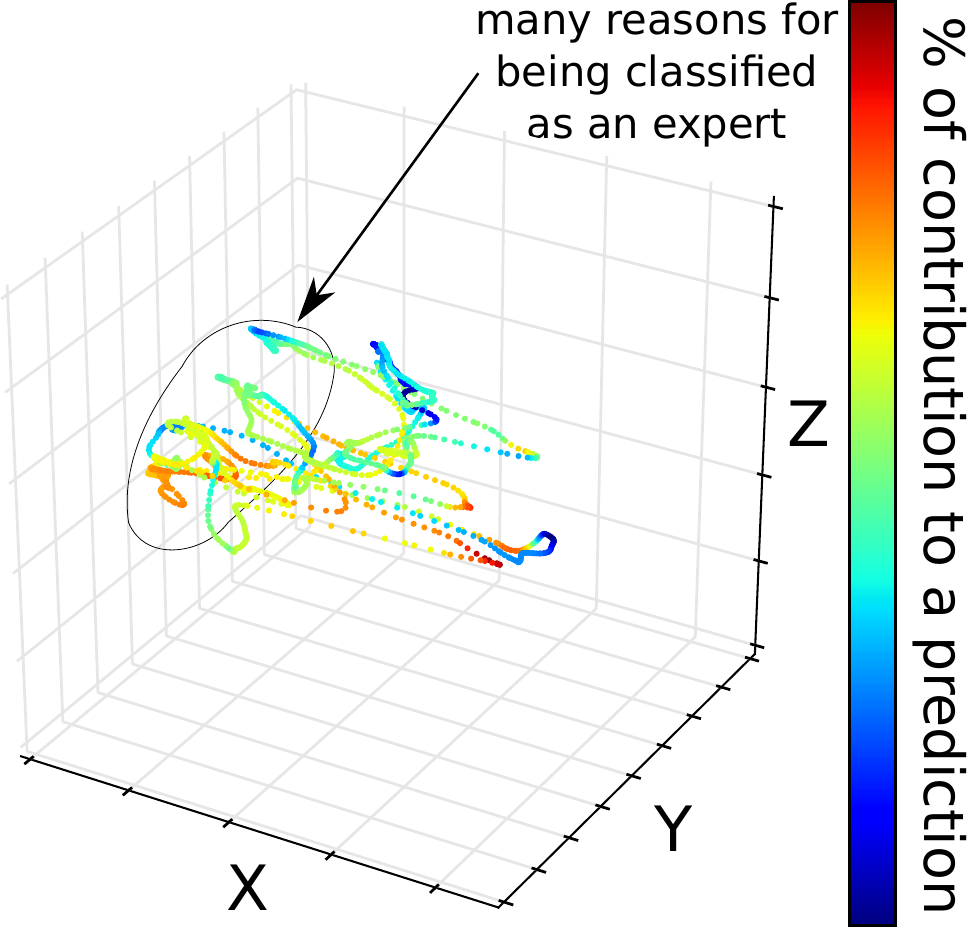}
      \label{sub:screenshot}}
 \hspace{.1cm}
    \subfloat[Suturing task for a novice]{
 \includegraphics[width=.35\linewidth]{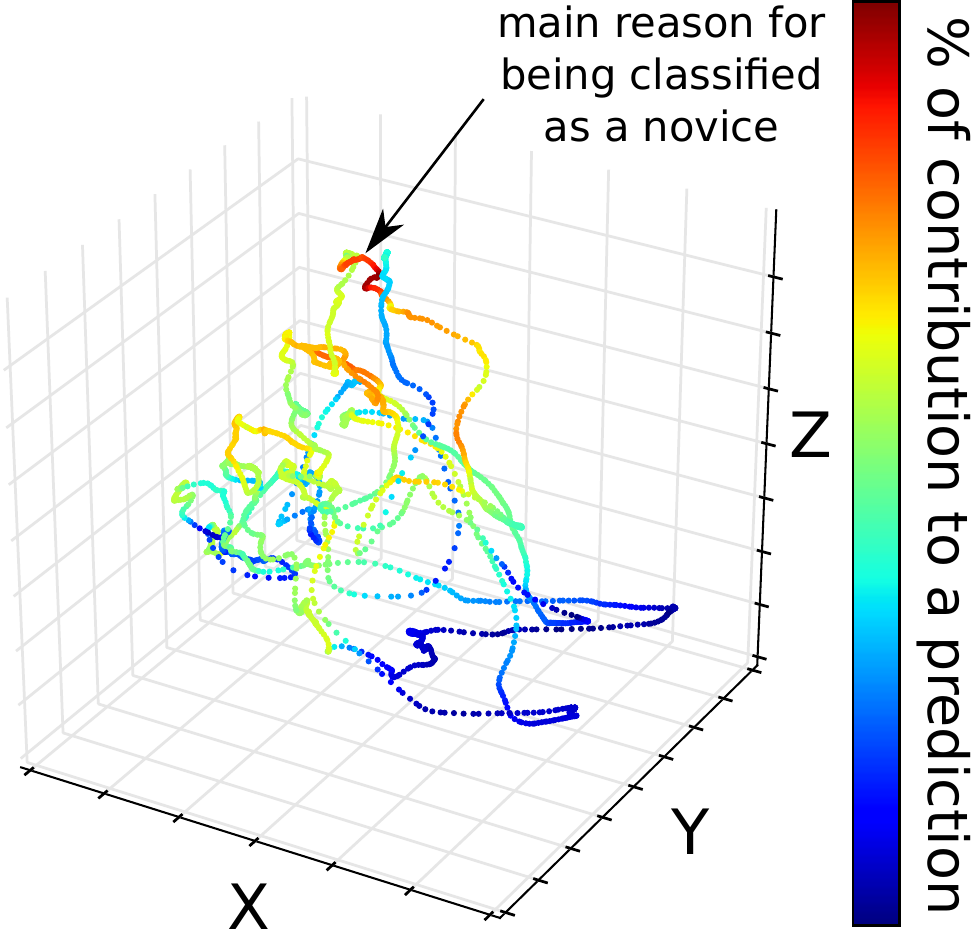}
      \label{sub:feedback}
      }
    \caption{Using class activation map (CAM) to provide explainable classification}
    \label{fig:trials}
\end{figure*}

\begin{figure*}
\centering
    \subfloat[Suture/needle handling]{
 \includegraphics[width=0.35\linewidth]{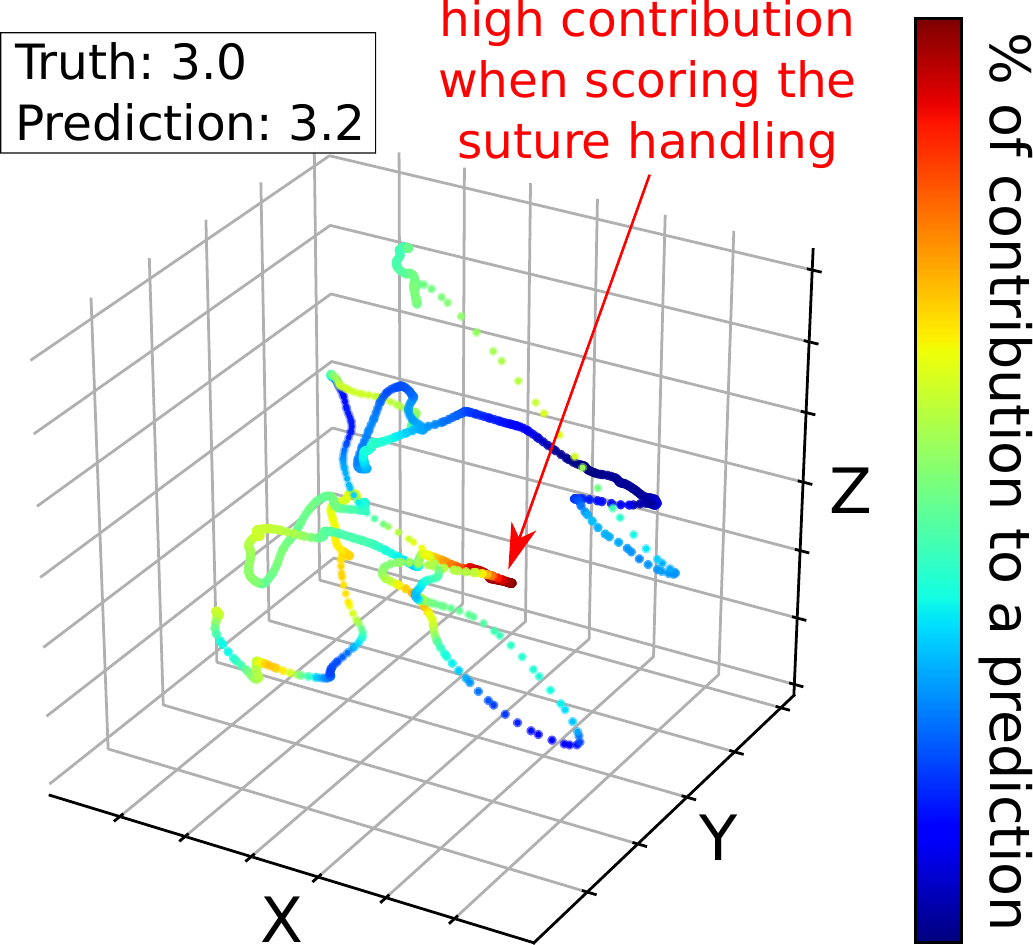}
      \label{sub-reg-2-kt-e002}}
      \subfloat[Quality of the final product]{
 \includegraphics[width=0.35\linewidth]{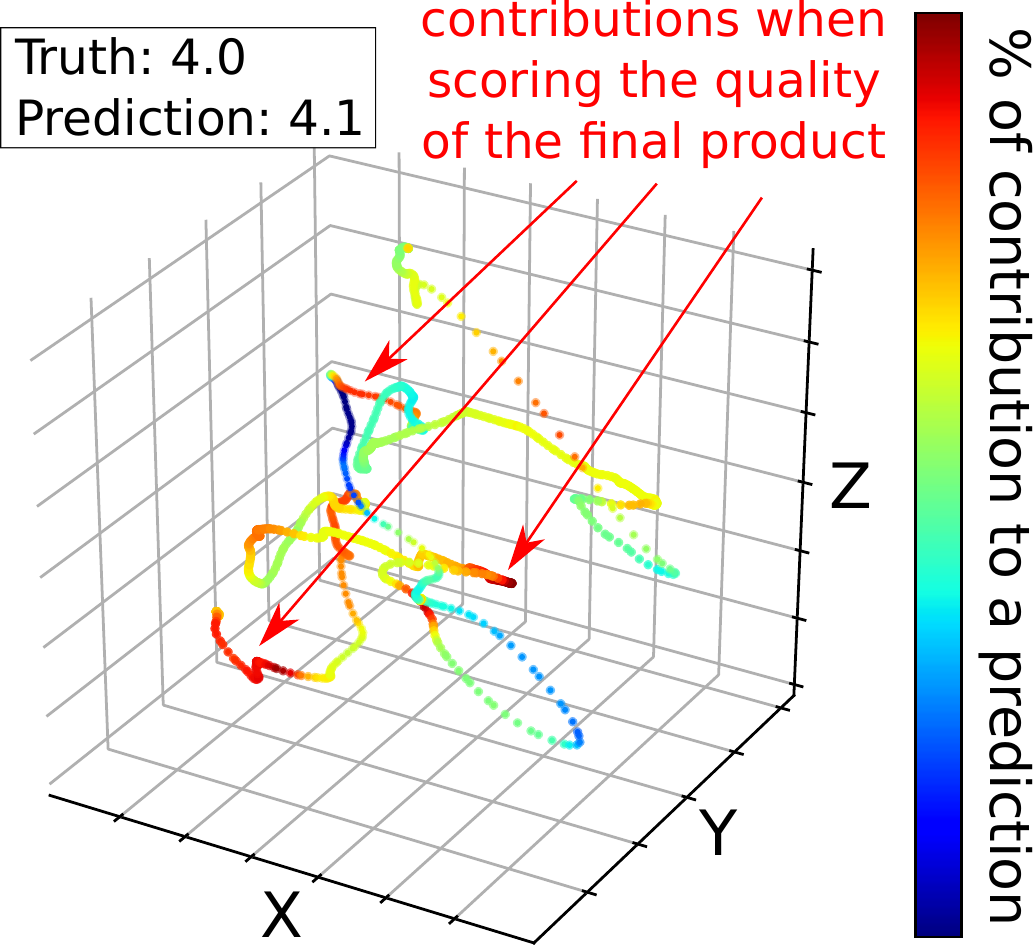}
      \label{sub-reg-6-kt-e002}}
    \caption{Feedback using the CAM on subject E's second knot-tying trial}
    \label{fig-cam-reg-kt}
\end{figure*} 

The CAM technique allows us to visualize which parts of the trial contributes the most to a skill classification. 
By localizing, for example, discriminative behaviors specific to a skill level, observers can start to understand motion patterns specific to certain class of surgeons. 
To further improve themselves (the novice surgeons), the model, using the CAM's result, can pinpoint to the trainees their good/bad motor behaviors. 
This would potentially enable novices to achieve greater performance and eventually become experts. 

By generating a heatmap from the CAM, we can see in \figurename~\ref{fig:trials} how it is indeed possible to visualize the feedback for the trainee. 
In fact, we examine a trial of an expert and novice surgeon: The expert's trajectory is illustrated in \figurename~\ref{sub:screenshot} while the novice's trajectory is depicted in \figurename~\ref{sub:feedback}.
In this example, we can see how the model was able to identify which motion (red subsequence) is the main reason for identifying a subject as a novice. 
Concretely, we can easily spot a pattern that is being recognized by the model when outputting the classification of subject H's skill level: The orange and red 3D subsequences correspond to same surgical gesture ``pulling suture'' and are exhibiting a high influence over the model's decision. 
This feedback could be used to explain to a young surgeon which movements are classifying him/her as a novice and which ones are classifying another subject as an expert. 
Thus ultimately, this sort of feedback could guide the novices into becoming experts.

After having shown how our \emph{classifier} can be interpreted to provide feedback to the trainees, we now present the result of applying the same visualization (based on the CAM algorithm) in order to explain the OSATS score prediction. 
Figure~\ref{fig-cam-reg-kt} depicts the trajectory with its associated heatmaps for subject E performing the second trial of the knot-tying task. 
Figure~\ref{sub-reg-2-kt-e002} and \ref{sub-reg-6-kt-e002} illustrates the trajectory's heatmap, respectively, for ``suture/needle handling'' and ``quality of the final product'' OSATS score predictions. 
At first glimpse, one can see how a prediction that requires focusing on the whole surgical trial leverages more than one region of the input surgery---this is depicted by the multiple red subsequences in \figurename~\ref{sub-reg-6-kt-e002}.
However, when outputting a rating for a specific task such as ``suture/needle handling''---the model is focusing on less parts of the input trajectory which is shown in \figurename~\ref{sub-reg-2-kt-e002}. 

\section{Limitations}
We would like to first highlight the fact that our feedback technique would benefit from an extended real use-case validation process, for example verifying with expert surgeons if indeed the model is able to detect the main reason for classifying a surgical skill.
In addition, the fact that we are performing only a LOSO setup means that a surgeon should be present in the training set in order to make a prediction. 
However, since only two experts exist in the dataset, this suggests that performing a leave-one-user-out setup would mean having one expert in the training set. 
This constitutes a huge problem originating from the limited dataset size. 
Therefore, we finally conclude that our approach should be validated on a larger dataset.

\section{Conclusion}\label{sec:conc}
In this paper, we proposed a novel deep learning-based method for surgical skills evaluation from kinematic data. 
We achieved state-of-the-art accuracy by designing a specific FCN, while providing explainability that justifies a certain skill evaluation, thus allowing us to mitigate the CNN's black-box effect.
Furthermore, by extending our architecture we were able to provide new state-of-the-art performance for predicting the OSATS score from the input kinematic time-series data. 
In the future, in order to compensate for the lack of labeled data, we aim at exploring several regularization techniques such as data augmentation and transfer learning~\cite{IsmailFawaz2018transfer} for time-series data. 

\begin{acknowledgements}
The authors would like to thank the creators of JIGSAWS, as well as NVIDIA Corporation for the GPU grant and the M\'esocentre of Strasbourg for providing access to the cluster.
The authors would also like to thank the MICCAI 2018 anonymous reviewers for their fruitful comments that helped us improve the quality of this manuscript.
\end{acknowledgements}

\begin{small}~

\noindent
\textbf{Compliance with Ethical Standards} \vspace{-0.25mm}\\

\noindent
\textbf{Conflict of interest}~ The authors declare that they have no conflict of interest. \\

\noindent
\textbf{Ethical approval}~ All procedures performed in studies involving human participants were in accordance with the ethical standards of the institutional and/or national research committee and with the 1964 Helsinki Declaration and its later amendments or comparable ethical standards. \\

\noindent \textbf{Informed consent}~ Informed consent was obtained from all individual participants included in the study. 
\end{small}

\begin{small}
\bibliographystyle{spbasic} 
\bibliography{biblio}
\end{small}


\end{document}